# REAL-TIME HUMAN-COMPUTER INTERACTION BASED ON FACE AND HAND GESTURE RECOGNITION

Reza Azad[1], Babak Azad[2], Nabil Belhaj Khalifa[3], Shahram Jamali[4]

[1]Department of Electrical and Computer Engineering, Shahid Rajaee Teacher Training University, Tehran, Iran
[2] Computer Engineering Department, University of Mohaghegh Ardabili, Ardabil, Iran
[3]Blaise Pascal University, Clermont Ferrand, France
[4]Associate professor, Department of Computer Engineering, University of Mohaghegh Ardabili, Ardabil, Iran

## ABSTRACT

*At the present time, hand gestures recognition system could be used as a more expected and useable approach for human computer interaction. Automatic hand gesture recognition system provides us a new tactic for interactive with the virtual environment. In this paper, a face and hand gesture recognition system which is able to control computer media player is offered. Hand gesture and human face are the key element to interact with the smart system. We used the face recognition scheme for viewer verification and the hand gesture recognition in mechanism of computer media player, for instance, volume down/up, next music and etc. In the proposed technique, first, the hand gesture and face location is extracted from the main image by combination of skin and cascade detector and then is sent to recognition stage. In recognition stage, first, the threshold condition is inspected then the extracted face and gesture will be recognized. In the result stage, the proposed technique is applied on the video dataset and the high precision ratio acquired. Additional the recommended hand gesture recognition method is applied on static American Sign Language (ASL) database and the correctness rate achieved nearby 99.40%. also the planned method could be used in gesture based computer games and virtual reality.*

## 1. INTRODUCTION

In the existing world, the communication with the intelligent devices has progressive to such a magnitude that as humans it has become essential and we cannot live without its capability. The new machinery has become so embedded into our regular lives that we use it to shop, work, interconnect and even interest our self [1]. It has been extensively supposed that the calculating, communiqué and presentation machineries progress extra, but the current systems may become a holdup in the effective operation of the existing information flow. For efficiently using of these systems, most computer applications need more and more communication. For that motive, human-computer interaction (HCI) has been a dynamic field of study in the last decades. Initially systems that are used for graphically HCI system are mouse and keyboards. Even if the innovation of the mouse and keyboard is a great development, there are still circumstances in which these devices are irreconcilable for HCI. This is principally the case for the communication with 3D objects.

The two points of freedom of the mouse could not suitably emulate the 3 dimensions of space. The use of hand gestures offers a smart and natural optional to these burdensome interface tools for human computer communication. With use of hands gesture recognition system can help people to interactive with computers in a more intuitive mode. Hand gesture recognition owns wide applications in sign language recognition [2], [3], computer games [4], virtual reality [5] and HCI systems [6]. There were numerous gesture recognition methods established for tracking and recognizing numerous hand gestures. Each one of them has their advantage and disadvantage. Wired technology is the oldest one, in which in order to connect or interface with the computer system, users need to tie up themselves with the help of wire. User cannot freely move in the room in wired technology as they connected with the computer system via wire and limited with the length of wire. The best example for the wired technology is the instrumented gloves -also called data gloves or electronics gloves. These electronics gloves have some sensors, and thanks to these sensors they provide information related to location of the hand, position orientation of finger etc. Output results of these data gloves are well but for the wide range common application using propose they are expensive through to being an electronic device [3].

After the Data gloves, the optical markers is appeared. The optical markers detect the location of hand or tips of fingers by projecting Infra-Red light and reflect this light on screen. These organizations also offer the worthy result but need a very complex configuration. nowadays some new methods have been proposed for hand gesture recognition, such as Image based systems which needs processing of image structures like texture, color etc. the approaches on optical markers are very luxurious and have very difficult configuration [3]. Also the technique based on image processing is weedy against under diverse illumination situation, color texture modifying, which leads to variations in observed outcomes. For improving the image processing based technique for hand gestures recognition scheme we planned current paper method. In this paper, we use only a video camera and a PC to progress a hand gesture based HCI system. Our methodology for HCI is contain of four stages, I. face and hand detection based on fusion of skin detection and cascade detector; II. extracting hand new position based on particle filter algorithm; III. measuring threshold condition based on hand new position and applying face & hand recognition stage; IV. Controlling smart device (in this paper we considered computer music player) by extracted information from third stage.

The rest of the paper is prepared as follows. In section two proposed methods is offered and in section three and four the practical result and conclusion are detailed respectively

## 2. PROPOSED METHOD FOR HCI

In the current paper, we advance a human-computer interaction scheme using a video camera to acquire images. Overall chart of the suggested method is depicted in Fig. 1. In the first stage of projected method, for hand and face detection, first of all we tag the regions of an image using skin colors, which performance as nominees for the face and hand. Next, connected components are exposed from these image regions. Third, we fixed a threshold for the connected components for removing noise, rejecting the zones which are tiny to be candidates for the face or hand. For the outstanding succeeded nominees, first, the length of each of these connected component is found and two area that has the highest length is nominated as face and hand candidate. Then between these two candidates the face location is selected by Viola jones detector.

### 2.1. Stage 1: Face & Hand Detection

In order to extract the face and hand from the images, a skin pixel finder, connected component (CC) generation and Viola jones detector has been implemented.

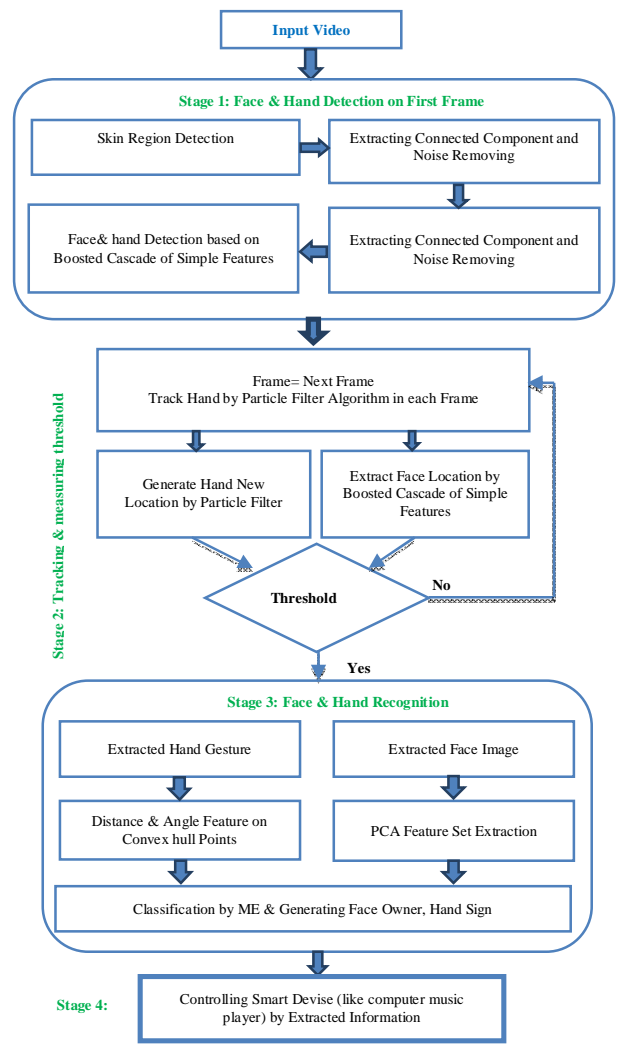

Figure 1. Overall illustration of the proposed method

### 2.1.1. Skin region detection

Skin region extracting is one of the greatest and primary step in face detection purpose [7]. In this respect, many approaches have been suggested to detect skin which showed high detection amount. In this paper we used our skin detection method that we cited earlier in [8]. Fig. 2(b) shows the result of this detector on the entry image.

### 2.1.2. Creating Connected Component by mathematical morphology

Mathematical morphology is one of the divisions of image processing that contends about structure and impression of abject in images [9]. After applying the skin detector, all possible skin areas are extracted and represented as white pixels. It's possible that the small noisy regions will be existing, but we undertake that the region related to the hand and faces are the biggest. Consequently we first destroy the noise by "disk" erosion. Those small noisy areas, wrongly discovered as hand skin, are typically skin-color like objects under certain light situation. The erosion procedure could successfully erase those small noisy areas. Nonetheless the real areas related to the hand and face may also shrink with considering the erosion procedure planned for noise. Accordingly we applied the dilation to strength the major discovered area, help to improve the anticipated hand and face detection. Meanwhile the hand and face are close areas; the holes are filled in this phase. After this progression, the two big areas extracted as face and hand candidate and rest of the area reduced. Fig. 2(c) shows the outcome.

### 2.1.3. Viola and Jones Detector for Face Detection

The Viola and Jones approached for object detection that introduced in 2001 has become one of the most common real-time frameworks. This method is basically a cascade of binary linear classifiers which are consequently applied to the sliding window input. For more detail see the [10]. In current paper we qualified the Viola-Jones framework with a lot of human faces, such as: natural, tilted, side view, darken and blurred with considering any distance and lighting condition. After training the classifier, at the running time the face area is nominated between two object that extracted by the last step. In Fig. 2(d) the blue square selected as face by applying this detector.

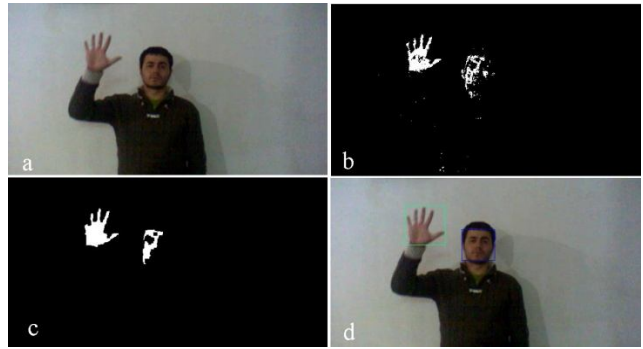

Figure 2.  (a): entrance image (b): detected skin region (c): result of mathematical morphology (d): detected face and hand region

### 2.2. Stage 2: Tracking & measuring threshold condition

In this stage, the new position of hand is extracted by particle filter algorithm. For this respect, we used the extracted hand image (from first frame), as input for particle filter algorithm. By applying this algorithm we achieved hand new position in each frame. Also we extracted face new position in each frame by Viola jones detector. After extracting face and hand new position we used (1) for measuring distance between these objects.

$$d = \sqrt{(x_2 - x_1)^2 - (y_2 - y_1)^2} \qquad (1)$$

In top relation $x_2$ and $y_2$ are the face center coordinates and $x_1$ and $y_1$ are the hand center coordinates in 2D space. By extracting the distance we used the following rule, also Fig. 3 shows the sample of threshold condition.

$$IF\ d \leq Threshold$$

$$then: goto\ Stage\ 3$$

$$Else: \text{Repeat this stage again on new frame.}$$

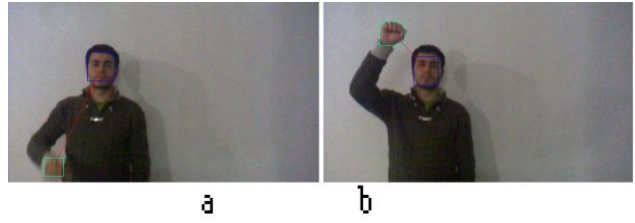

Figure 3. (a): distance between face and hand is bigger than threshold (b): distance between face and hand is smaller than threshold

## 2.3. Stage 3: Face & Hand Recognition

### 2.3.1. Hand gesture feature set

In our scheme we calculated features with use of angel and distance of intersections points of the images as follows: First the minimum rectangle containing the hand gesture location is extracted. Then for improving the accuracy and removing the dependency of features to size, position we transformed each image to a standard size of 100×100 pixels. We selected this normalized value based of several experiments. Then we take out the outline point of images. Finally we used the Graham algorithm ($O(n\,log\,n)$) for extracting the intersections points. Fig. 4 shows the sample of these intersections points.

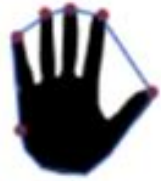

Figure 4. Intersection points of hand gesture

Then we extracted angle and distance feature set by (2) and (3) respectively:

$$(a_b) = \theta(b_i, b_{i+1}) \qquad i = 1,2,3,\dots \quad (2)$$

$$(y_b) = \sqrt{\left(b_{i+1_x} - b_{i_x}\right)^2 - \left(b_{i+1_y} - b_{i_y}\right)^2} \quad i = 1,2,3,\dots \quad (3)$$

In top relation $a_b$ is angel feature set for any hand gesture and $\theta(b_i, b_{i+1})$ is angle of two intersection points to image horizontal level and $y_b$ is distance feature set for any hand gesture and $i = 1,2,3,\dots$ is intersection points of hand gesture. By using this instruction we achieved 6 features for each gesture. As second feature set we mined the Fourier descriptors. Fourier descriptor is a moral descriptor of the outlines, and is rotational, scale and translation invariance. Fourier descriptors are achieved by computing Fourier factors of the arrangement of gestures edge point. The technique telling the gesture feature has nothing to do with the initial point in boundary and recognizes hand gesture fast. By using a specific point on the boundary of subdivision gesture as the initial point, coordinate structure of the boundary is reached by counter-clockwise as mentioned in (4):

$$z(k) = [x(k), y(k)], k = 0,1,\dots, n-1 \quad (4)$$

And the plural formula is as (5):

$$p(l) = x(l) + jy(l), (l = 0,1,\dots,n-1), j = \sqrt{1-} \quad (5)$$

2D problem will be transformed into a 1D problem. The border of 1D Discrete Fourier coefficient sequence is demarcated as (6):

$$z(k) = \frac{1}{n}\sum_{l=0}^{n-1} p(l) \exp\left(-j\frac{2\pi lk}{n}\right), k = 0,1,\dots,n-1 \quad (6)$$

The Fourier descriptors of example gestures is shown in Fig. 5 and attained as Table.1.

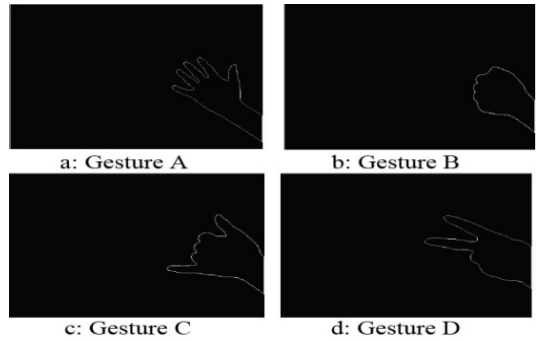

a: Gesture A    b: Gesture B

c: Gesture C    d: Gesture D

Figure 5. Contour of gesture. (a): gesture A (b): gesture B (c): gesture C (d): gesture D

Table 1. The Fourier descriptors of the hand gestures

| Hand Gesture Class | Fourier Feature set | | | | | | |
|---|---|---|---|---|---|---|---|
| A | 1.000 | 0.3529 | 0.1970 | 0.0587 | 0.0095 | 0.0975 | 0.1143 |
| B | 1.000 | 0.1514 | 0.0046 | 0.0067 | 0.0062 | 0.0028 | 0.0010 |
| C | 1.000 | 22.513 | 0.7957 | 0.4742 | 1.5620 | 0.0111 | 0.0827 |
| D | 1.000 | 0.0513 | 0.0010 | 0.0272 | 0.0121 | 0.0073 | 0.0001 |

### 2.3.2. Face feature set

There are many approaches for face recognition field, such as [11], [12]. In this part we used Principal Component Analysis (PCA) technique for feature extraction. The PCA was offered by Karl Pearson and Harold Hotelling to transform a set of feasibly correlated variables into a reduced set of uncorrelated variables [13]. The notion is that a high-dimensional database is frequently designated by correlated variables and for this reasons only a few significant dimensions account for maximum of the information. The PCA approaches find the directions with the highest variance in the data, called principal components.

### 2.3.3. Classification bay Mixture of Experts

In order to combining classifiers, there are two main strategies: selection and fusion. In classifier selection, every member is assigned to learn a part the feature space, whereas in classifier fusion, it is supposed that on the whole feature space, each ensemble member is trained. The mixture of experts (ME) is one of the most popular methods of classifier selection, which originally proposed by Jacobs et al. [14]. Expert combination actually is a classic strategy that has been broadly used in various problem solving tasks [15-17]. A group of single with diverse and complementary skills tackles a task jointly such that a performance higher than any single individual can make is achieved via integrating the strengths of individuals [18].

Architecture of the ME is composed of N local experts and also, for defining the outputs expert weights conditioned on the input, there is a gating network. In our proposed method, there is a hidden layer for every expert $i$- a multi-layer perceptron (MLP) neural network-, which computes an output $Oi$ as a function of the input stimuli vector $x$, and also, there are output layers, a sigmoid activation function, and a set of hidden weights. We suppose that in a different area of the input space each expert specializes. A weight $gi$ assigned to each of the expert's output, $O_i$ by the gating network. Also, the $g = \{g_1, g_2, \dots, g_N\}$ determined as a function of the input vector x, and a set of parameters determined as weights of its hidden; and hole of output layers and a sigmoid activation function, determined by the gating network. Where the expert i can generate the desired output y, every element gi of g can be interpreted as estimates of the prior probability. The MLP neural network and softmax nonlinear operator are the gating networks two constitutive layers. Thus the gating network computes the output vector of the MLP layer of the gating network- $\tau = \{\tau_1, \tau_2, \dots, \tau_N\}$-, then applies the softmax function to get (7):

$$g_i = \frac{\exp(\tau_i)}{\sum_{j=1}^{N} \exp(\tau_i)}, i = 1, 2, \dots, N \ (7)$$

Here $g_i$s are non-negative and sum to 1, and N is the number of expert networks. The final mixed output of the entire network is (8):

$$T = \sum_i o_i g_i, i = 1, 2, \dots, N \ (8)$$

By the error Back Propagation (BP) algorithm, the weights of MLPs are learned. For the gating network and each expert i, the weights are updated according to the (9):

$$\Delta_{w_i} = \eta_e h_i (y - O_i) \big(O_i(1 - O_i)\big) V_i^T,$$
$$\Delta_{\omega_i} = \eta_e h_i W_i^T (y - O_i) \big(O_i(1 - O_i)\big) \big(V_i(1 - O_i)\big) x,$$
$$\Delta_\xi = \eta_g (h - g) \big(\tau(1 - \tau)\big) \vartheta^T,$$
$$\Delta_\varsigma = \eta_g \xi^T (h - g) \big(\tau(1 - \tau)\big) \vartheta^T (1 - \vartheta) x, (9)$$

For the expert and the gating networks, the rate of learning shown by $\eta_e$ and $\eta_g$. Weight matrices of input to hidden and hidden to output layer, shown by the $\omega$ and $w$ respectively. $\zeta$ and $\xi$ are the weight matrices of hidden to output layer and input to hidden, respectively, for the gating network. $V_i^T$ and $\vartheta^T$ are the transpose of $v_i$ and $\vartheta$, the output matrices of the hidden layer of expert and gating networks, respectively. In the above formulas $h = \{h_1, h_2, \dots, h_N\}$ is a vector such that each $h_i$ is an estimate of the posterior probability that expert i can generate the desired outputy, and is computed as (10):

$$h_i = \frac{g_i \exp\left(\frac{-1}{2}(y - o_i)^T (y - o_i)\right)}{\sum_j g_i \exp\left(\frac{-1}{2}(y - o_j)^T (y - o_j)\right)} \ (10)$$

## 2.4. Stage 4: Controlling smart device

After extracting information (face owner and hand sign) from last stage, we used this information for controlling computer music player by Matlab functions. Also this stage could be used as application in: smart device controlling, smart TV, robots, computer game and etc.

## 3. PRACTICAL AND IMPLEMENTATION RESULT

Our mentioned method has been done on Intel Core i3-2330M CPU, 2.20 GHz with 2 GB RAM with use of Matlab software. In Fig. 6 the face of worked systems is shown. Data achievement, mixture of experts' configuration and performance of the proposed system are labels in the next subsections.

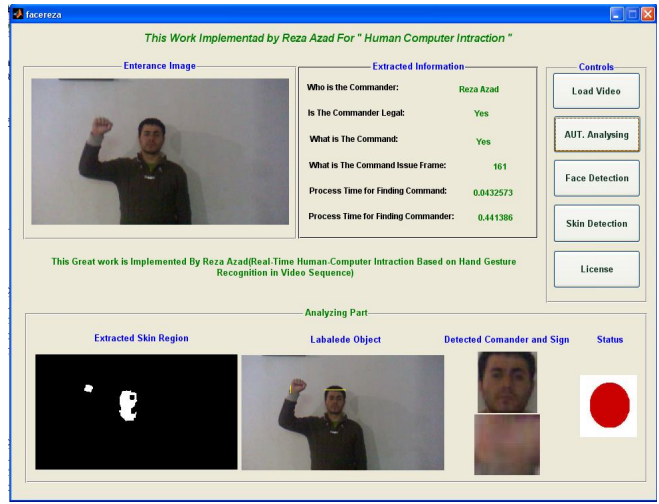

Figure 6. Human computer interaction system implemented in Matlab software

### 3.1. Data Achievement

The database which we used for human hand gesture recognition while moving comprises five types of gesture obtained from five persons with dissimilar scene. The video sequence has been arranged using a fixed place 3 Mega Pixel Nokia 5233 while the person is moving toward it. Fig. 7 shows examples of these gestures.

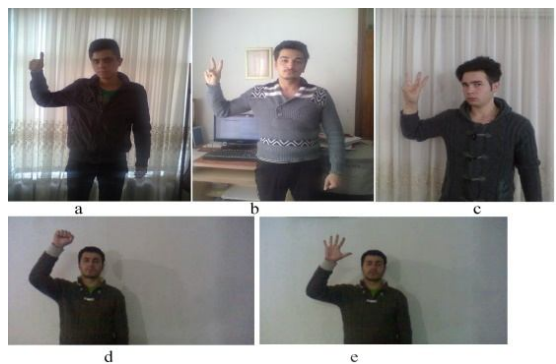

Figure 7. Hand gesture signs (a): Stop music (b): Play music (c): Next music (d): Volume up (e): Volume down

### 3.2. Mixture of Experts Configuration

As proposed earlier, our neural network scheme contains several MLP neural networks that perform the experts' role and they are mixed through the mixture of experts' methodology. The training set comprises the intersections point's features and Fourier descriptors features of the 50% images of train set and the other 50% images are used for the testing. Subsequently the input data are in the 17 dimensional space the topology of the planned network should have 17 nodes in the input layer and also since the number of gestures is 5, then the quantity of the nodes used in the output layer must be 5 (every node represents one hand gesture). Consequently, our proposed neural networks topologies are differ only in the quantity of the hidden layer nodes. A lot of configurations of the network by altering the complexity of the experts or the quantity of experts are tested in implementing the neural network and also different values are used parameters configuration. In whole the experimental result the gating learning rate was equal to 0.4 and the amount of its hidden nodes was 40 nodes, the expert learning degree was 0.9, and the network trained by 300 epochs. We tasted our system on different number of hidden nodes of experts and the outcomes are detailed in table 2.

Table 2. The recognition degree based on diverse number of hidden nodes and experts on the hand gesture database

| Quantity of Nodes | Hand Gesture Database | | |
|---|---|---|---|
| | Two Experts | Three Experts | Four Experts |
| 10 nodes | 79.50% | 81% | 80% |
| 15 nodes | 90.23% | 96% | 96% |
| 20 nodes | 94.75% | 99.20% | 97.53% |
| 30 nodes | 95% | 98.05% | 96.86% |

As it is clear from the table 2, the accuracy rate for the schemes having 10 nodes in their hidden layer is proportionately low. With expanding the quantity of hidden nodes from 10 to 15 the accuracy rate improves meaningfully, and in the two- experts system with 20 hidden nodes the accuracy rate of 94.75% proof that the quantity of experts was deficient that can't divide the input space appropriately. In the four-expert organization with the similar number of hidden nodes matching to the system with three-experts there are numerous free parameters that makes the network too complex to get a well result than 3-experts. Thus we used the network with three-experts and 20 numbers of hidden nodes as our classifier because it distributes the input space in the greatest way and establish a balance of the amount of experts and hidden nodes.

## 3.3. Performance Evaluation of the Proposed Method

The experimental outcomes proofed that the proposed system has a robust recognition level in detecting and recognition human computer interaction technique. Table 3 characterizes the experimental results. Na, NCR and AR respectively refer to number of gesture in videos, number of correct recognition and the accuracy rate.

Table 3. The recognition rate for various gestures in video sequence

| Hand Gesture | NA | NCR | AR |
|---|---|---|---|
| G1 (Stop music) | 50 | 50 | 100% |
| G2 (Play music) | 50 | 49 | 98% |
| G3 (Next music) | 50 | 50 | 100% |
| G4 (Volume up) | 50 | 49 | 98% |
| G5 (Volume down) | 50 | 50 | 100% |

Furthermore, for well understanding of the wrong classification on outcomes we have depicted the confusion matrix of the classifier yield in Table 4. Notice that G1-G4 respectively representative of these hand sign gesture: stop music, play music, next music, volume up and volume down.

Table 4. Confusion matrix of the proposed hand gesture recognition system

| Classified As | | | | | |
|---|---|---|---|---|---|
| | G1 | G2 | G3 | G4 | G5 |
| G1 | 50 | 0 | 0 | 0 | 0 |
| G2 | 0 | 49 | 0 | 1 | 0 |
| G3 | 0 | 0 | 50 | 0 | 0 |
| G4 | 0 | 0 | 0 | 49 | 1 |
| G5 | 0 | 0 | 0 | 0 | 50 |

Further for additional experimental enquiry, we had tested the above mention system on [3] static hand gesture database. Table 5 shows the accuracy level for each hand gesture.

Table 5. Recognition rate of the proposed method for each static hand gesture

| Hand Gesture class | Number of Input Images | Recognized Image | | Precision Rate of Each Technique | |
|---|---|---|---|---|---|
| | | Technique [3] | proposed method | Technique [3] | proposed method |
| A | 21 | 20 | 21 | 95.23% | 100% |
| B | 21 | 21 | 21 | 100% | 100% |
| C | 21 | 21 | 21 | 100% | 100% |
| D | 21 | 21 | 21 | 100% | 100% |
| E | 21 | 19 | 20 | 90.47% | 95.23% |
| F | 21 | 21 | 21 | 100% | 100% |
| G | 21 | 21 | 21 | 100% | 100% |
| H | 21 | 21 | 21 | 100% | 100% |
| I | 21 | 21 | 21 | 100% | 100% |
| K | 21 | 21 | 21 | 100% | 100% |
| L | 21 | 21 | 21 | 100% | 100% |
| M | 21 | 19 | 20 | 90.47% | 95.23% |
| N | 21 | 21 | 21 | 100% | 100% |
| O | 21 | 21 | 21 | 100% | 100% |
| P | 21 | 21 | 21 | 100% | 100% |
| Q | 21 | 21 | 21 | 100% | 100% |
| R | 21 | 21 | 21 | 100% | 100% |
| S | 21 | 20 | 21 | 95.23% | 100% |
| T | 21 | 21 | 21 | 100% | 100% |
| U | 21 | 21 | 21 | 100% | 100% |
| V | 21 | 21 | 20 | 100% | 95.23% |
| W | 21 | 21 | 21 | 100% | 100% |
| X | 21 | 21 | 21 | 100% | 100% |
| Y | 21 | 21 | 21 | 100% | 100% |
| Total | 504 | 498 | 501 | 98.80% | 99.40% |

Great detection degree shows the quality of proposed methodology to use in every applications, which are desired a HCI. Also we attained 100% accuracy in face recognition stage.

# 4. CONCLUSIONS

We offered a human-computer interaction (HCI) scheme using a PC and a video camera established on face and hand gesture recognition. A face recognition stage was used for viewer verification and the hand gesture recognition stage for monitoring computer media player. In our suggested method, first, we extracted the hand and face location from the main image by combination of skin discovery and Viola Jones detector. After extracting face and hand we used particle filter algorithm and threshold condition for applying recognition stage. Finally in the recognition stage the feature set for face and hand gesture extracted respectively and recognized by the mixture of experts. In the result stage, our proposed method is tested on the video dataset and we achieved proximally 99.20% accuracy rate. Auxiliary we applied the mentioned algorithm on static American Sign Language (ASL) database and we obtained 99.40% correctness ratio.

## ACKNOWLEDGEMENTS

This research is supported by the Blaise Pascal University and the SRTTU, Tehran, Iran (No.22970060-9).

## Authors

**Reza Azad** obtained his B.Sc. degree with honor in computer software engineering from SRTTU in 2014. He is IEEE & IEEE conference reviewer Member. Awarded as best student in 2013 and 2014 by the SRTTU and awarded as best researcher in 2013 by the SRTTU. He achieved fourth place in Iranian university entering exam. In addition he's a member of Iranian elites. He has a lot of scientific papers in international journal and conferences, such as IEEE, Springer and etc. his interested research are artificial intelligence and computer vision.

**Babak Azad** is a researcher from Islamic Azad University. He achieved a lot of awards and publication on scientific papers in international journals and conferences, during his B.Sc. education. His most interest topics are machine learning and network.

**Nabil BELHAJ KHALIFA** is a student of Master Research at Blaise Pascal University, France. Actually, he is an intern at LIMOS, a research laboratory in Clermont-Ferrand in France, in order to achieve his M.Sc. degree in computer science in field of computer vision. He obtained his Engineering diploma in computer science from ISSATSo university in Tunisia in 2012. And  before, he get his B.Sc. degree in computer software from ISSATSo, Tunisia. His research interests include image processing, computer vision, computer graphics, machine learning and artificial neural networks.

**Shahram Jamali** is currently an Associate Professor in Mohaghegh Ardabili University, Ardebil, Iran.  He achieved his Ph.D degree in Architecture of Computer Systems in 2008 from Iran University of Science & Technology, Tehran, Iran. He has more than 100 scientific papers in international journals and conferences, such as IEEE, Elsevier, Springer and etc. His research topics are Network security and soft computing.